\documentclass[11pt, a4paper]{article}
\usepackage{coling2018}
\usepackage{times}
\usepackage{latexsym}

\usepackage{adjustbox}
\usepackage{url}
\usepackage{amsmath}
\usepackage{amssymb}
\usepackage{bm}
\usepackage{graphicx}
\usepackage{mathtools}
\usepackage{multirow,makecell}
\usepackage{caption}
\usepackage{subcaption}
\usepackage{booktabs}
\usepackage{tikz}
\usetikzlibrary{automata,positioning}
\usepackage{color}
\usepackage{longtable}
\definecolor{black}{rgb}{0,0,0}

\newcommand\numberthis{\addtocounter{equation}{1}\tag{\theequation}}
\newcommand{\norm}[1]{\left\lVert#1\right\rVert}

\title{Retrofitting Distributional Embeddings to Knowledge Graphs\\ with Functional Relations}
\author{Benjamin J. Lengerich \\
  Carnegie Mellon University \\
  5000 Forbes Avenue \\
  Pittsburgh, PA 15213 \\
  {\tt blengeri@cs.cmu.edu} \\\And
  Andrew L.~Maas \and Christopher Potts \\
  Stanford University, Roam Analytics \\
  195 East 4th Avenue \\
  San Mateo, CA 94401 \\
  {\tt {\{amaas, cgpotts\}}@roaminsight.com} \\
  }

\date{}

\begin{document}
\maketitle
\begin{abstract}
Knowledge graphs are a versatile framework to encode richly structured data relationships, but it can be challenging to combine these graphs with unstructured data. Methods for retrofitting pre-trained entity representations to the structure of a knowledge graph typically assume that entities are embedded in a connected space and that relations imply similarity. However, useful knowledge graphs often contain diverse entities and relations (with potentially disjoint underlying corpora) which do not accord with these assumptions. To overcome these limitations, we present \textit{Functional Retrofitting}, a framework that generalizes current retrofitting methods by explicitly modeling pairwise relations. Our framework can directly incorporate a variety of pairwise penalty functions previously developed for knowledge graph completion. Further, it allows users to encode, learn, and extract information about relation semantics. We present both linear and neural instantiations of the framework. Functional Retrofitting significantly outperforms existing retrofitting methods on complex knowledge graphs and loses no accuracy on simpler graphs (in which relations do imply similarity). Finally, we demonstrate the utility of the framework by predicting new drug--disease treatment pairs in a large, complex health knowledge graph.
\end{abstract}

\blfootnote{This work is licensed under a Creative Commons Attribution 4.0 International License. License
details: \url{http://creativecommons.org/licenses/by/4.0/}}

\section{Introduction}
Distributional representations of concepts are often easy to obtain from unstructured data sets, but they tend to provide only a blurry picture of the relationships that exist between concepts. In contrast, knowledge graphs directly encode this relational information, but it can be difficult to summarize the graph structure in a single representation for each entity.

To combine the advantages of distributional and relational data, \newcite{Faruqui-etal:2015} propose to \textit{retrofit} embeddings learned from distributional data to the structure of a knowledge graph. Their method first learns entity representations based solely on distributional data and then applies a retrofitting step to update the representations based on the structure of a knowledge graph. This modular approach conveniently separates the distributional data and entity representation learning from the knowledge graph and retrofitting model, allowing one to flexibly combine, reuse, and adapt existing representations to new tasks.

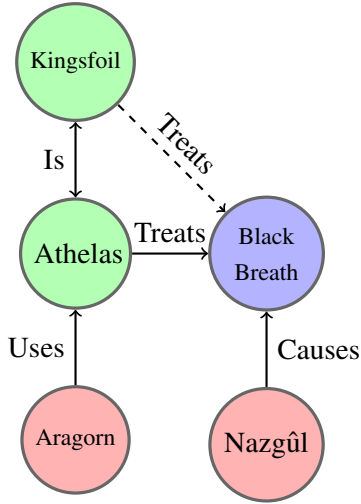
\begin{figure}[htb]
    \centering

    \begin{tikzpicture}[
    drug/.style={circle, draw=black!60, fill=green!30, very thick, minimum size=12mm,align=center},
    disease/.style={circle, draw=black!60, fill=blue!30, very thick, minimum size=12mm,align=center},
    concept/.style={circle, draw=black!60, fill=red!30, very thick, minimum size=12mm,align=center},
    ]
    \node[drug]         (athelas)           [text width=11mm]                   {Athelas};
    \node[drug]         (kingsfoil)         [above=of athelas, text width=12mm] {\footnotesize Kingsfoil};
    \node[disease]      (blackbreath)       [right=of athelas, text width=9mm]  {\small Black Breath};
    \node[concept]      (nazgul)            [below=of blackbreath]              {Nazg\^{u}l};
    \node[concept]      (aragorn)           [below=of athelas]                  {\small Aragorn};
     
    \draw[thick,<->] (kingsfoil) -- (athelas) node[midway,left] {Is};
    \draw[thick,->] (athelas) -- (blackbreath) node[midway,sloped,above] {Treats};
    \draw[thick,dashed,->] (kingsfoil) -- (blackbreath) node[midway,sloped,above] {Treats};
    \draw[thick,->] (aragorn) -- (athelas) node[midway,left] {Uses};
    \draw[thick,->] (nazgul) -- (blackbreath) node[midway,right] {Causes};
    \end{tikzpicture}
    \caption{Toy knowledge graph with diverse relation types that connect treatments (green), diseases (blue), and persons (red) by known (solid) and unknown (dashed) relations. 
    Traditional methods, which assume that all relations imply similarity, would retrofit Aragorn and Nazg\^{u}l toward similar embeddings.}
    \label{fig:toy_graph}
\end{figure}

However, a core assumption of \newcite{Faruqui-etal:2015}'s retrofitting model is that connected entities should have similar embeddings. This assumption often fails to hold in large, complex knowledge graphs, for a variety of reasons. First, subgraphs of a knowledge graph often contain distinct classes of entities that are most naturally embedded in disconnected vector spaces. In the extreme case, the representations for these entities might derive from very different underlying data sets. For example, in a health knowledge graph, the subgraphs containing diseases and drugs should be allowed to form disjoint vector spaces, and we might want to derive the initial representations from radically different data sets. Second, many knowledge graphs contain diverse relationships whose semantics are different from -- perhaps even in conflict with -- similarity.  For instance, in the knowledge graph in Figure~\ref{fig:toy_graph}, the model of \newcite{Faruqui-etal:2015} would model embeddings $\boldsymbol{q_{Aragorn}} \approx \boldsymbol{q_{Athelas}} \approx \boldsymbol{q_{Black Breath}} \approx \boldsymbol{q_{Nazg\hat{u}l}}$, which is problematic as Aragorn is not semantically similar to a Nazg\^{u}l (they are enemies).

To address these limitations, we present \emph{Functional Retrofitting}, a retrofitting framework that explicitly models pairwise relations as functions. The framework supports a wide range of different instantiations, from simple linear relational functions to complex multilayer neural ones. Here, we evaluate both linear and neural instantiations of Functional Retrofitting on a variety of diverse knowledge graphs. For benchmarking against existing approaches, we use FrameNet and WordNet. We then move into the medical domain, where knowledge graphs play an important role in knowledge accumulation and discovery. 
These experiments show that even simple instantiations of Functional Retrofitting significantly outperform baselines on knowledge graphs with semantically complex relations and sacrifice no accuracy on graphs where \newcite{Faruqui-etal:2015}'s assumptions about similarity do hold. Finally, we use the model to identify promising new disease targets for existing drugs.

Code which implements Functional Retrofitting is available at \url{https://github.com/roaminsight/roamresearch}.

\section{Notation}
A knowledge graph $\mathcal{G}$ is composed of a set of vertices $\mathcal{V}$, a set of relation types $\mathcal{R}$, and a set of edges $\mathcal{E}$ where each edge $e\in \mathcal{E}$ is a tuple $(i,j,r)$ in which the relationship $r\in \mathcal{R}$ holds between vertices $i\in \mathcal{V}$ and $j\in \mathcal{V}$. Our goal is to learn a set of representations $\mathcal{Q} = \{\boldsymbol{q_i} : i \in \mathcal{V}\}$ which contain the information encoded in both the distributional data and the knowledge graph structure, and can be used for downstream analysis. Throughout this paper, we use $a$ to refer to a scalar, $\boldsymbol{a}$ to refer to a vector, and $\boldsymbol{A}$ to refer to a matrix or tensor.

\section{Related Work}
\label{sec:related_work}
Here we are interested in \textit{post-hoc} retrofitting methods, which adjust entity embeddings to fit the structure of a previously unseen knowledge graph.

\subsection{Retrofitting Models}
The primary introduction of retrofitting was \newcite{Faruqui-etal:2015}, in which the authors showed the value of retrofitting semantic embeddings according to minimization of the weighted least squares problem
\begin{displaymath}
\Psi_{\mathcal{G}}(\mathcal{Q}) = \sum_{i\in\mathcal{V}}\alpha_i\norm{\boldsymbol{q_i} - \boldsymbol{\hat{q}_i}}^2 + \sum_{(i,j,r)\in \mathcal{E}}\beta_{ij}\norm{\boldsymbol{q_i} - \boldsymbol{q_j}}^2 \numberthis{}
\label{eq:faruqui}
\end{displaymath}
where $\hat{\mathcal{Q}} = \{\boldsymbol{\hat{q}_i}: i \in \mathcal{V}\}$ is the embedding learned from the distributional data and $\alpha_i$, $\beta_{ij}$ set the relative weighting of each type of data. When $\alpha_i=1$ and $\beta_{ij} = \frac{1}{degree(i)}$, this model assigns equal weight to the distributional data and the structure of the knowledge graph.

More recently, \newcite{hamilton2017inductive} presented GraphSAGE, a two-step method which learns both an aggregation function $f: \mathbb{R}^{d\times n}\rightarrow \mathbb{R}^{k}$, to condense the representations of neighbors into a single point, and an update function $g: \mathbb{R}^{k+d}\rightarrow \mathbb{R}^d$, to combine the aggregation with a central vertex. Here, $d$ is the embedding dimensionality, $k$ is the aggregation dimensionality, and $n>0$ is the number of neighbors for each vertex. Note that $k>d$ is permitted, allowing for aggregation by concatenation. While this method is extremely effective for learning representations on simple knowledge graphs, it is not formulated for knowledge graphs with multiple types of relations. Furthermore, when the representation of a relation is known \textit{a priori}, it can be useful to explicitly set the penalty function (e.g., \newcite{mrkvsic2016counter} use hand-crafted functions to effectively model antonymy and synonymy). By aggregating neighbors into a point estimate before calculating relationship likelihoods, GraphSAGE makes it difficult to encode, learn, or extract the representation of a pairwise relation.

In a similar vein, \newcite{faruqui2016morpho} developed a graph-based semi-supervised learning method to expand morpho-syntactic lexicons from seed sets. Though the task is different from the retrofitting task we consider here, the performance and scalability of their method demonstrate the utility of directly encoding pairwise relations as message-passing functions.

\subsection{Relational Penalty Functions}
Our new Functional Retrofitting framework models each relation via a penalty function $f_r: \mathbb{R}^{d_i+d_j}\rightarrow \mathbb{R}_{\ge 0}$ acting on a pair of entities $(i,j)$ with embedding dimensionalities $d_i$ and $d_j$, respectively. By explicitly modeling relations between pairs of entities, Functional Retrofitting supports the use of a wide array of scoring functions that have previously been developed for knowledge graph completion. Here, we present a brief review of such scoring functions; for an extensive review, see \cite{nickel2016review}.

TransE \cite{bordes2013translating} uses additive relations in which the penalty function $f_r(\boldsymbol{q_i}, \boldsymbol{q_j}) = \norm{\boldsymbol{q_i} + \boldsymbol{a_r} - \boldsymbol{q_j}}^2_2$ is low iff $(i, j, r) \in \mathcal{E}$. The simple Unstructured Model \cite{bordes2012joint} was proposed as a na{\"i}ve version of TransE that assigns all $\boldsymbol{a_r} = \boldsymbol{0}$, leading to the penalty function $f_r(\boldsymbol{q_i}, \boldsymbol{q_j}) = \norm{\boldsymbol{q_i}-\boldsymbol{q_j}}_2^2$. This is the underlying penalty function of \cite{Faruqui-etal:2015}. It cannot consider multiple types of relations. In addition, while it models  1-to-1 relations well, it struggles to model multivalued relations.

TransH \cite{wang2014knowledge} was proposed to address this limitation by using multiple representations for a single entity via relation hyperplanes. For a relation $r$, TransH models the relation as a vector $\boldsymbol{a_r}$ on a hyperplane defined by normal vector $\boldsymbol{w_r}$. For a triple $(i, j, r) \in \mathcal{E}$, the entity embeddings $\boldsymbol{q_i}$ and $\boldsymbol{q_j}$ are first projected to the hyperplane of $\boldsymbol{w_r}$. By constraining $\norm{\boldsymbol{w_r}}_2 = 1$, we have the penalty function $f_r(\boldsymbol{q_i}, \boldsymbol{q_j}) = \norm{g_r(\boldsymbol{q_i}) + \boldsymbol{a_r} - g_r(\boldsymbol{q_j})}^2_2$ where $g_r(\boldsymbol{x}) = \boldsymbol{x} - \boldsymbol{w_r}^T\boldsymbol{x}\boldsymbol{w_r}$.

TransR \cite{lin2015learning} embeds relations in a separate space from entities by a relation-specific matrix $\boldsymbol{M_r}\in \mathbb{R}^{d \times k}$ that projects from entity space to relation space and a shared relation vector $\boldsymbol{a}\in \mathbb{R}^{k}$ that translates in relation space by $f_r(\boldsymbol{q_i}, \boldsymbol{q_j}) = \norm{\boldsymbol{q_i}\boldsymbol{M_r} + \boldsymbol{a} - \boldsymbol{q_j}\boldsymbol{M_r}}_2^2$. We use this model as the inspiration for our linear penalty function.

The Neural Tensor Network (NTN; \newcite{socher2013reasoning}) defines a score function $f_r(\boldsymbol{q_i}, \boldsymbol{q_j}) = \boldsymbol{u_r}^Tg(\boldsymbol{q_i}^T\boldsymbol{M_r}\boldsymbol{q_j} + \boldsymbol{M_{r,1}}\boldsymbol{q_i} + \boldsymbol{M_{r,2}}\boldsymbol{q_j} + \boldsymbol{b_r})$ 
where $\boldsymbol{u_r}$ is a relation-specific linear layer, $g: \mathbb{R}^{k}\rightarrow \mathbb{R}^{k}$ is the tanh operation applied element-wise, $\boldsymbol{M_r} \in \mathbb{R}^{d\times d\times k}$ is a 3-way tensor, and $\boldsymbol{M_{r,1}},\boldsymbol{M_{r,2}} \in \mathbb{R}^{k\times d}$ are weight matrices. All of these models can be directly incorporated in our Functional Retrofitting framework.

\section{Functional Retrofitting}
We propose the framework of Functional Retrofitting (FR) to incorporate a set $\mathcal{F}$ of relation-specific penalty functions $f_r:\mathbb{R}^{d_i+d_j}\rightarrow \mathbb{R}_{\ge 0}$ which penalizes embeddings of entities $i,j$ with dimensionalities $d_i,d_j$, respectively. 
This gives the complete minimization:
\begin{align*}
\Psi_{\mathcal{G}}(\mathcal{Q}; \mathcal{F}) = &\sum_{i\in\mathcal{Q}}\alpha_i||\boldsymbol{q_i} - \boldsymbol{\hat{q}_i}||^2 + \sum_{(i,j,r)\in \mathcal{E}}\beta_{i,j,r}f_{r}(\boldsymbol{q_i}, \boldsymbol{q_j}) - \sum_{(i,j,r)\in \mathcal{E}^-}\beta_{i,j,r}f_{r}(\boldsymbol{q_i}, \boldsymbol{q_j}) + \sum_{r\in \mathcal{R}}\rho_{\lambda}(f_r)\numberthis{}
\end{align*}
where $\boldsymbol{\hat{q}_i}$ is observed from distributional data, $\alpha_i$ and $\beta_{i,j,r}$ set the relative strengths of the distributional data and the knowledge graph structure, and $\rho$
regularizes $f_r$ with strength set by $\lambda$. $\mathcal{E}^-$ is the \emph{negative space} of the knowledge graph, a set of edges that are not annotated in the knowledge graph. FR uses $\mathcal{E}^-$ to penalize relations that are implied by the representations but not annotated in the graph. To populate $\mathcal{E}^-$, we sample a single negative edge $(i,j',r)$ with the same outgoing vertex for each true edge $(i,j,r)\in\mathcal{E}$. The user can calibrate trust in the completeness of the knowledge graph via the $\beta$ hyperparameter.

In contrast to prior retrofitting work, FR explicitly encodes directed relations. This allows the model to fit graphs which contain diverse relation types and entities embedded in disconnected vector spaces. Here, we compare the performance of two instantiations of FR -- one with all linear relations and one with all neural relations -- and show that even these simple models provide significant performance improvements. In practice, we recommend that users select relation-specific functions in accordance with the semantics of their graph's relations.



\subsection{Linear Relations}
We implement a linear relational penalty function $f_r(\boldsymbol{q_i}, \boldsymbol{q_j})=\norm{\boldsymbol{A_r}\boldsymbol{q_j} + \boldsymbol{b_r} - \boldsymbol{q_i}}^2 $
with $\ell_2$ regularization for minimization of:
\begin{align*}
\Psi_{\mathcal{G}}(\mathcal{Q}; \mathcal{F}) &= \sum_{i=1}^n\alpha_i\norm{\boldsymbol{q_i} - \boldsymbol{\hat{q}_i}}^2 + \sum_{(i,j,r)\in \mathcal{E}}\beta_{i,j,r}\norm{\boldsymbol{A_r}\boldsymbol{q_j} + \boldsymbol{b_r} - \boldsymbol{q_i}}^2 \\
&- \sum_{(i,j,r)\in \mathcal{E}^-}\beta_{i,j,r}\norm{\boldsymbol{A_r}\boldsymbol{q_j} + \boldsymbol{b_r} - \boldsymbol{q_i}}^2 + \lambda\sum_{r\in\mathcal{R}}||\boldsymbol{A_r}||^2\numberthis{}
\label{eq:linear_loss}
\end{align*}
\subsubsection*{Identity Relations}
\newcite{Faruqui-etal:2015}'s model is a special case of this formulation in which
\begin{align*}
\boldsymbol{A_r} &= \boldsymbol{I},\quad\boldsymbol{b_r} =\boldsymbol{0}\quad\forall r, \quad\quad \beta_{i,j,r} = \begin{cases} \frac{1}{\mathit{degree}(i)} & (i,j,r)\in \mathcal{E} \\
0 & (i,j,r) \in \mathcal{E}^-
\end{cases}
\end{align*}
Throughout the remainder of this paper, we refer to this baseline model as the ``FR-Identity" retrofitting method.

\subsubsection*{Initialization}
We initialize embeddings as those learned from distributional data and relations to imply similarity:
\begin{align*}
    &\boldsymbol{A_r} = \boldsymbol{I},\quad \boldsymbol{b_r} = \boldsymbol{0} \quad,
       \quad \alpha_i = \begin{cases}
        0 & \boldsymbol{\hat{q}_i} = \boldsymbol{0} \\
        \alpha & \boldsymbol{\hat{q}_i} \neq \boldsymbol{0} \end{cases},
        \quad \beta_{i,j,r} = \begin{cases}
        \frac{\beta^+}{d_r(i)} & (i,j,r) \in \mathcal{E}\\
        \frac{\beta^-}{d_r(i)} & (i,j,r) \in \mathcal{E}^- \end{cases}
\end{align*}
where $d_r(i)$ is the out-degree of vertex $i$ for relation type $r$, $\alpha$ is a hyperparameter to trade off distributional data against structural data, and $\beta$ sets the trust in completeness of the knowledge graph structure. In our experiments, we use $\beta^+=1,~\beta^-=0$ for straightforward comparison with the method of \newcite{Faruqui-etal:2015} and optimize $\alpha$ by cross-validation. Given prior knowledge about the semantic meaning of relations, we could initialize relations to respect these meanings (e.g., antonymy could be represented by $\boldsymbol{A_r} = -\boldsymbol{I}$).

\subsubsection*{Learning Procedure}
We optimize this model by block optimization. Conveniently, we have closed-form solutions where the partial derivatives of Eq. \ref{eq:linear_loss} equal $0$:
\begin{align*}
\boldsymbol{b_r} &= \frac{\sum\limits_{(i,j)}(-1)^{I_{\{(i,j,r)\notin\mathcal{E}\}}}\beta_{i,j,r}(\boldsymbol{A_rq_j}-\boldsymbol{q_i})} {\sum\limits_{(i,j)}(-1)^{I_{\{(i,j,r)\notin\mathcal{E}\}}}\beta_{i,j,r}} \numberthis{} \\
\boldsymbol{\tilde{A}_r} &= \boldsymbol{U}\boldsymbol{V}^{-1} \numberthis{}\\
\boldsymbol{U} &= \sum_{(i,j):(i,j,r)\in \mathcal{E}}\beta_{i,j,r}(\boldsymbol{q_i} - \boldsymbol{b_r})\boldsymbol{q_j}^T - \sum_{(i,j):(i,j,r)\in \mathcal{E}^-}\beta_{i,j,r}(\boldsymbol{q_i} - \boldsymbol{b_r})\boldsymbol{q_j}^T \numberthis{}\\
\boldsymbol{V} &= \sum_{(i,j): (i,j,r)\in \mathcal{E}}\beta_{i,j,r}\boldsymbol{q_j}\boldsymbol{q_j}^T - \sum_{(i,j): (i,j,r)\in \mathcal{E}^-}\beta_{i,j,r}\boldsymbol{q_j}\boldsymbol{q_j}^T + \lambda \boldsymbol{I} \numberthis{}
\end{align*}
Constraining $\boldsymbol{A_r}$ to be orthogonal by $\boldsymbol{A_r} = \boldsymbol{\tilde{A}_r}(\boldsymbol{\tilde{A}_r}^T\boldsymbol{\tilde{A}_r})^{-1/2}$, we have $\boldsymbol{q_i} = \frac{\boldsymbol{a_i}}{b_i}$ where 
\begin{align*}
\boldsymbol{a_i} = \alpha_i\boldsymbol{\hat{q}_i} &+ \sum\limits_{(j,r):(i,j,r)\in \mathcal{E}}\beta_{i,j,r}(\boldsymbol{A_r}\boldsymbol{q_j}+\boldsymbol{b_r}) + \sum\limits_{(j,r):(j,i,r)\in \mathcal{E}}\beta_{j,i,r}\boldsymbol{A_r}^T(\boldsymbol{q_j} - \boldsymbol{b_r}) \\
 &- \sum\limits_{(j,r):(i,j,r)\in \mathcal{E}^-}\beta_{i,j,r}(\boldsymbol{A_r}\boldsymbol{q_j}+\boldsymbol{b_r}) - \sum\limits_{(j,r):(j,i,r)\in \mathcal{E}^-}\beta_{j,i,r}\boldsymbol{A_r}^T(\boldsymbol{q_j} - \boldsymbol{b_r}) \numberthis{}\\
 b_i = \alpha_i &+ \sum\limits_{(j,r):(i,j,r)\in \mathcal{E}}\beta_{i,j,r} + \sum\limits_{(j,r):(j,i,r)\in \mathcal{E}}\beta_{j,i,r} - \sum\limits_{(j,r):(i,j,r)\in \mathcal{E}^-}\beta_{i,j,r} - \sum\limits_{(j,r):(j,i,r)\in \mathcal{E}^-}\beta_{j,i,r} \numberthis{}
\label{eq:linear_orthogonal}
\end{align*}

\subsection{Neural Relations}
We also instantiate FR with a neural penalty function $f_r(\boldsymbol{q_i}, \boldsymbol{q_j}) = \sigma(\boldsymbol{q_i}^T\boldsymbol{A_r}\boldsymbol{q_j})$ 
where $\sigma$ is the element-wise tanh operation, $\boldsymbol{A_r} \in \mathbb{R}^{d_i \times d_j}$, again with $\ell_2$ regularization.
We initialize weights in a similar manner as for the linear relations and update via stochastic gradient descent. In our experiments, we use $\beta^+=\beta^-=1$, and sample the same number of non-neighbors as true neighbors.

\section{Experiments}
\label{sec:experiments}
We test FR on four knowledge graphs. The first two are standard lexical knowledge graphs (FrameNet, WordNet) in which FR significantly improves retrofitting quality on complex graphs and loses no accuracy on simple graphs. The final two graphs are large healthcare ontologies (SNOMED-CT, Roam Health Knowledge Graph), which demonstrate the scalability of the framework and the utility of the new embeddings.

For each graph, we successively evaluate link prediction accuracy after retrofitting to links of other relation types. Specifically, for each relation type $r\in\mathcal{R}$, we retrofit to $\mathcal{G}_{\backslash r} = (\mathcal{V},\mathcal{E}_{\backslash r})$ where $\mathcal{E}_{\backslash r}=\{(i,j,r'):(i,j,r')\in \mathcal{E} \text{, } r'\neq r\}$ is the set of edges with all relations of type $r$ removed. After retrofitting, we train a Random Forest classifier to predict the presence of relation $r$ between entities $i$ and $j$ (with 70\% of vertices selected as training examples and the remainder reserved for testing).  To have balanced class labels, we sample an equivalent number of non-edges, $\mathcal{E}^-_{r} = \{(i,j,r):(i,j,r)\notin \mathcal{E}\}$ with $|\mathcal{E}^-_{r}| = |\mathcal{E}|$ and
$|\{j : (i,j,r) \in \mathcal{E}^-_{r} \}| = |\{j : (i,j,r) \in \mathcal{E}\}|~\forall~i$.
Thus, the random baseline classification rate is set to $50\%$. Other baselines are the embeddings built from distributional data and the retrofitting method of \newcite{Faruqui-etal:2015}, denoted as ``None" and ``FR-Identity", respectively. 

\subsection{FrameNet}
\label{sec:framenet_description}
FrameNet \cite{baker1998berkeley,fillmore2003background} is a linguistic knowledge graph containing information about lexical and predicate argument semantics of the English language.
FrameNet contains two distinct entity classes: \textit{frames} and \textit{lexical units}, where a \textit{frame} is a meaning and a \textit{lexical unit} is a single meaning for a word. 
To create a graph from FrameNet, we connect lexical unit $i$ to frame $j$ if $i$ occurs in $j$. We denote this relation as ``Frame", and its inverse ``Lexical unit".
Finally, we connect frames by the structure of FrameNet (Table~\ref{tab:framenet_structure}). Distributional embeddings are from the Google News pre-trained Word2Vec model \cite{mikolov2013distributed}
; the counts of each entity type that were also found in the distributional corpus are shown in Table~\ref{tab:framenet_structure}.

\subsubsection*{Results}
As seen in Table \ref{tab:framenet}, the representations learned by FR-Linear and FR-Neural are significantly more useful for link prediction than those of the baseline methods. 

\begin{table*}[htp]
    \centering
    \setlength{\tabcolsep}{5pt}
    \begin{adjustbox}{center}
    \begin{tabular}{l c c c c c}
        \toprule
        \multirow{2}{20mm}{\makecell{\bf{Retrofitting}\\\bf{Model}}} & \bf{{\footnotesize `Inheritance'}} & \bf{{\footnotesize `Using'}} & \bf{{\footnotesize `Reframing Mapping'}} & \bf{{\footnotesize `Subframe'}} & \bf{{\footnotesize `Perspective On'}}   \\
                            &  (2132/992)         & (1552/668)    & (544/312)                & (356/168)       & (336/148)                \\
        \midrule
        None                & $87.58\pm1.04$                & $88.59\pm1.93$        & $85.60\pm1.80$         & $91.24\pm0.86$     & $89.59\pm3.25$                        \\
        FR-Identity            & $90.79\pm0.69$                & $87.87\pm1.48$        & $87.02\pm0.63$         & $94.50\pm1.70$     & $\underline{94.24\pm1.02}$                   \\
        FR-Linear              & $\boldsymbol{92.92\pm0.16}$  & $\underline{92.04\pm1.45}$   & $\underline{89.37\pm2.45}$ & $\underline{94.65\pm1.05}$     & $\boldsymbol{94.73\pm1.12}$              \\
        FR-Neural              & $\underline{92.46\pm0.67}$     & $\boldsymbol{92.54\pm1.45}$        & $\boldsymbol{89.57\pm0.70}$         & $\boldsymbol{95.65\pm2.21}$ & $94.04\pm0.58$     
    \\
    \bottomrule
    \end{tabular}
    \end{adjustbox}
    \begin{adjustbox}{center}
    \begin{tabular}{l c c c c}
        \\
        \toprule
        \multirow{2}{2cm}{\makecell{\bf{Retrofitting}\\\bf{Model}}} & \bf{`Precedes'}  & \bf{`See Also'} & \bf{`Causative Of'} & \bf{`Inchoative Of'}    \\
                            & (220/136)          &  (268/76)       & (204/36)             & (60/16)                     \\
        \midrule
        None                & $\underline{87.30\pm4.33}$       & $85.11\pm3.20$      & $86.11\pm6.00$             & $\underline{82.50\pm14.29}$                          \\
        FR-Identity            & $85.26\pm4.46$       & $83.81\pm2.14$               & $84.49\pm8.72$             & $78.33\pm20.14$                          \\
        FR-Linear              & $87.00\pm2.18$ & $\underline{91.93\pm1.06}$   & $\underline{92.09\pm6.34}$   & $\underline{82.50\pm14.29}$             \\
        FR-Neural              & $\boldsymbol{89.16\pm5.60}$  & $\boldsymbol{93.25\pm1.79}$    & $\boldsymbol{94.33\pm4.68}$        & $\boldsymbol{85.00\pm7.07}$                 
    \\
    \bottomrule
    \end{tabular}
    \end{adjustbox}
    \caption{Retrofitting to FrameNet. Reported values are mean and standard deviation of the link prediction accuracies over three experiments. The number of edges used for (training/testing) is shown below each edge type.
    }
    \label{tab:framenet}
\end{table*}

\subsection{WordNet}
\label{sec:wordnet_description}
WordNet \cite{miller1995wordnet,fellbaum_2005} is a lexical database consisting of words (lemmas) which are grouped into unordered sets of synonyms (synsets). To examine the performance of FR on knowledge graphs which predominately satisfy the assumptions of \newcite{Faruqui-etal:2015}, we extract a simple knowledge graph of lemmas and the connections between these lemmas that are annotated in WordNet. These connections are dominated by hypernymy and hyponymy (Table~\ref{tab:wordnet_structure}), which correlate with similarity, so we expect the baseline retrofitting method to perform well.

\subsubsection*{Results}
As seen in Table~\ref{tab:wordnet}, the increased flexibility of the FR framework does not degrade embedding quality even when this extra flexibility is not intuitively necessary. Here, we evaluate standard lexical metrics for word embeddings: word similarity and syntatic relations. For word similiarity tasks, the evaluation metric is the Spearman correlation between predicted and annotated similarities; for syntatic relation, the evaluation metric is the mean cosine similarity between the learned representation of the correct answer and the prediction by the vector offset method \cite{mikolov2013linguistic}. In contrast to our other experiments, here the only stochastic behavior is due to stochastic gradient descent training, not sampling of evaluation samples. Even though the WordNet knowledge graph largely satisfies the assumptions of the na{\"i}ve retrofitting model, the flexible FR framework achieves sustained improvements for both word similarity datasets (WordSim-353; \newcite{finkelstein2001placing}, Mturk-771\footnote{\url{http://www2.mta.ac.il/~gideon/mturk771.html}}, and MTurk-287
and syntatic relations (Google Analogy Test Set\footnote{\url{http://download.tensorflow.org/data/questions-words.txt}}).

\begin{table*}[htp]
    \centering
    \setlength{\tabcolsep}{8pt}
    \begin{adjustbox}{center}
    \begin{tabular}{l c c c c}
    \toprule
        \multirow{2}{2cm}{\makecell{\bf{Retrofitting}\\\bf{Model}}} & \multicolumn{3}{c}{\bf{Word Similarity}} & \multicolumn{1}{c}{\bf{Syntactic Relation}}\\
                      & WordSim-353   & MTurk-771            & MTurk-287 & Google Analogy \\
        \midrule
        None & $0.512$                & $0.538$              & $0.671$               & $0.772$ \\
        FR-Identity & $0.512$            & $0.532$              & $0.664$               & $0.774$ \\
        FR-Linear & $\boldsymbol{0.542}$ & $\boldsymbol{0.562}$ & $\boldsymbol{0.679}$  & $\boldsymbol{0.793}$ \\
        FR-Neural & $\underline{0.516\pm0.001}$         & $\underline{0.543\pm0.001}$              & $\underline{0.676\pm0.001}$               & $\underline{0.784\pm0.000}$
    \\
    \bottomrule
    \end{tabular}
    \end{adjustbox}
    \caption{Retrofitting to WordNet. Reported values are Spearman correlations for the word similarity tasks and mean cosine similarity for the syntatic relation task. These are deterministic evaluations, so the only source of stochasticity is the optimization of the FR-Neural model.}
    \label{tab:wordnet}
\end{table*}

\subsection{SNOMED-CT}
\label{sec:snomed}
SNOMED-CT is an ontology of clinical healthcare terms and concepts including diseases, treatments, anatomical terms, and many other types of entities. From the publicly available SNOMED-CT knowledge graph,\footnote{\url{https://www.nlm.nih.gov/healthit/snomedct/index.html}} we extracted 327,001 entities and 3,809,639 edges of 169 different types (Table~\ref{tab:snomed_structure}). 
To create distributional embeddings, we first link each SNOMED-CT concept to a set of Wikipedia articles by indexing the associated search terms in WikiData.\footnote{\url{https://dumps.wikimedia.org/wikidatawiki/entities/}} We aggregate each article set by the method of \newcite{arora2016simple}, which performs TF-IDF weighted aggregation of pre-trained term embeddings to create sophisticated distributional embeddings of SNOMED-CT concepts. This creates a single 300-dimensional vector for each entity.

\subsubsection*{Results}
As the SNOMED-CT ontology is dominated by synonymy-like relations, we expect the simple retrofitting methods to perform well. Nevertheless, we see minimal loss in link prediction performance by using the more flexible FR framework (Table \ref{tab:snomed}). Our implementation supports the use of different function classes to represent different relation types; in practice, we recommend that users select function classes in accordance with relation semantics.

\begin{table*}[htp]
	\centering
	\setlength{\tabcolsep}{8pt}
    \begin{adjustbox}{center}
    \begin{tabular}{l c c c c}
        \toprule
        \multirow{2}{2cm}{\makecell{\bf{Retrofitting}\\\bf{Model}}} &  \bf{`Has Finding Site'} & \bf{`Has Pathological Process'} & \bf{`Due to'} & \bf{`Cause of'}\\
                      & (113748/49070)          & (19318/8124)                    & (5042/2042)   & (1166/376) \\
        \midrule 
        None          & $95.26\pm0.01$                    & $98.79\pm0.07$                       & $91.47\pm0.88$              & $79.61\pm1.27$      \\
        FR-Identity      &  $\underline{95.25\pm0.11}$    & $\underline{99.09\pm0.11}$                   & $\boldsymbol{94.69\pm0.61}$ & $\boldsymbol{86.67\pm1.27}$ \\
        FR-Linear        & $\boldsymbol{95.35\pm0.01}$       & $\boldsymbol{99.35\pm0.01}$      & $\underline{93.50\pm0.46}$              & $\underline{80.82\pm0.49}$      \\
        FR-Neural        & $95.22\pm0.00$                        & $98.97\pm0.22$                                & $91.70\pm0.15$              & $80.29\pm0.80$      \\
    \bottomrule
    \end{tabular}
    \end{adjustbox}
    \caption{Retrofitting to SNOMED-CT. Reported values are mean and standard deviation of the link prediction accuracies over three experiments. The number of edges used for (training/testing) is shown below each edge type.}
    \label{tab:snomed}
\end{table*}

\subsection{Roam Health Knowledge Graph}
Finally, we investigate the utility of FR in the Roam Health Knowledge Graph (RHKG). The RHKG is a rich picture of the world of healthcare, with connections into numerous data sources: diverse medical ontologies, provider profiles and networks, product approvals and recalls, population health statistics, academic publications, financial data, clinical trial summaries and statistics, and many others. As of June 2, 2017 the RHKG contains 209,053,294 vertices, 1,021,163,726 edges, and 6,231,287,999 attributes. Here, we build an instance of the RHKG using only public data sources involving drugs and diseases. The structure of this knowledge graph is summarized in Table~\ref{tab:rhkg_structure}. 
In total, we select 48,649 disease--disease relations, 227,051 drug--drug relations, and 13,667 drug--disease relations used for retrofitting. 
A disjoint set of 11,306 drug--disease relations is reserved for evaluation.

In the RHKG, as in many industrial knowledge graphs, different distributional corpora are available for each type of entity. First, we mine 2.9M clinical texts for co-occurrence counts in physician notes. After counting co-occurrences, we perform a pointwise mutual information transform and $\ell_2$ row normalization to generate embeddings for each entity. For drug embeddings, we supplement these embeddings with physician prescription habits. We extract prescription counts for each of 808,020 providers in the 2013 Centers for Medicare \& Medicaid (CMS) dataset\footnote{\url{https://www.cms.gov/Research-Statistics-Data-and-Systems/Statistics-Trends-and-Reports/Medicare-Provider-Charge-Data/Part-D-Prescriber.html}} 
and 837,629 providers in the 2014 CMS dataset. By aggregating prescriptions counts across provider specialty, we produce 201-dimensional distributional embeddings for each drug. Finally, we retrofit these distributional embeddings to the structure of the knowledge graph (excluding `Treats' edges reserved for evaluation).

\subsubsection*{Results}
\begin{figure*}[ht]
	\centering
	\includegraphics[width=\textwidth]{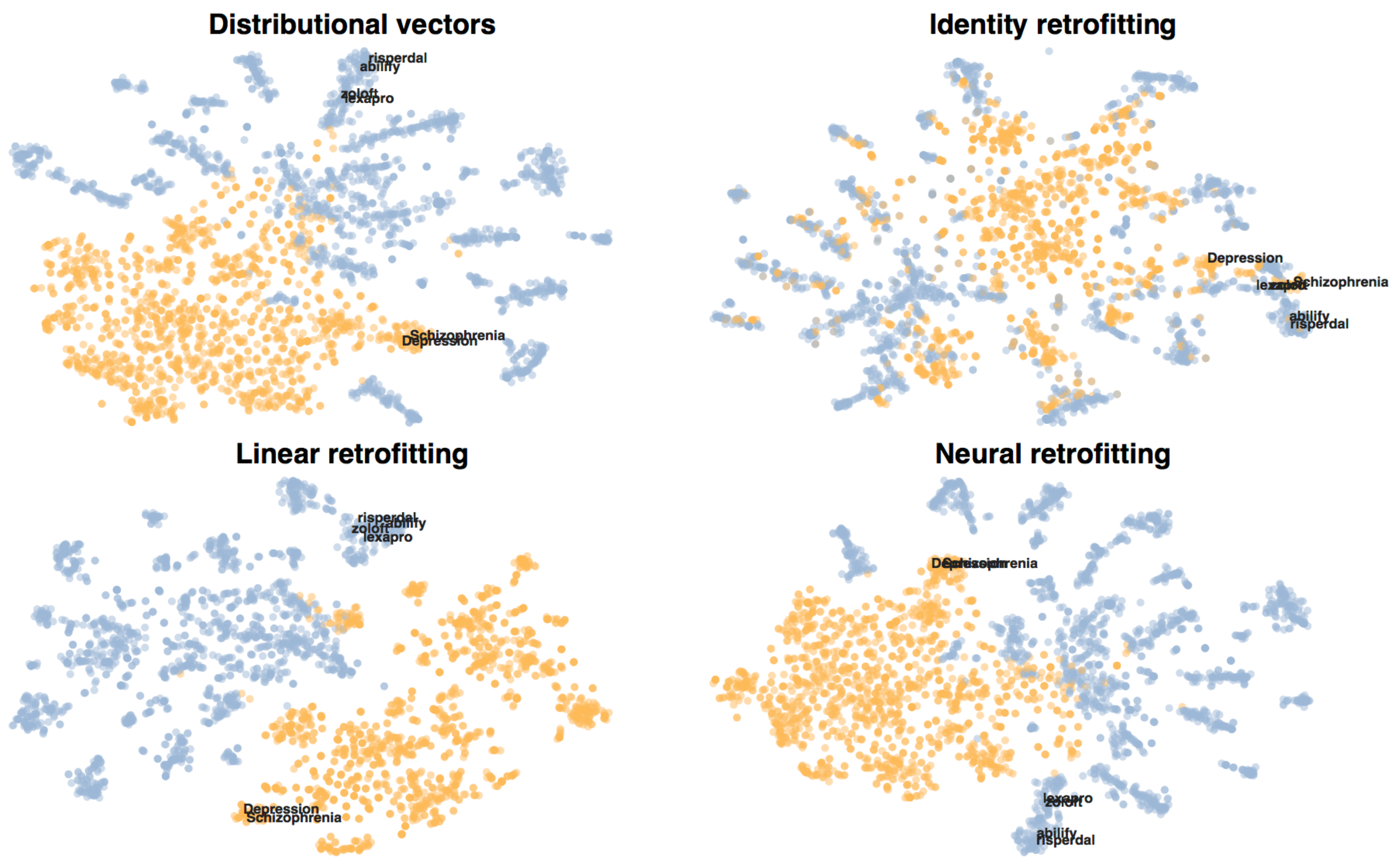}
	\caption{t-SNE Projections of the retrofitted embeddings of the drugs (blue) and diseases (orange) in the Roam Health Knowledge Graph, with selected annotations reflecting the `Treats' relation. The distributional space strongly separates the two kinds of entity because their representations were learned in different ways. Identity retrofitting blurs this basic semantic distinction in order to make diseases and drugs in `Treats' relations more similar. As Table \ref{tab:rhkg} shows, the FR models achieve this same unification, but they need not distort the basic drug/disease distinction to do it.}
	\label{fig:rhkg}
\end{figure*}

As shown in Table \ref{tab:rhkg}, the FR framework significantly improves prediction of `Treats' relations. We hypothesize that this is due to the separable nature of the graph; as seen in Figure~\ref{fig:rhkg}, the FR retrofitting framework can learn Disease and Drug subgraphs that are nearly separable. In contrast, Identity retrofitting generates a single connected space and distorts the embeddings.

\begin{table*}[ht]
\centering
{\footnotesize
\begin{adjustbox}{center}
\begin{tabular}{l l l r r}
\toprule
\multirow{2}{2cm}{\makecell{\bf{Retrofitting}\\\bf{Model}}} & \bf{Drug} & \bf{Disease Target} & \multirow{2}{9mm}{\bf{Model}\\\bf{Score}} & \bf{Plausible}\\
 & & & & \\
\midrule
\multirow{5}{*}{\makecell{None}}   & Naproxen               & Ankylosing Spondylitis                      & $0.98$ & Y\\
                                   & Latanoprost            & Superficial injury of ankle, foot and toes  & $0.96$ & N \\
                                   & Pulmicort              & Psoriasis, unspecified                      & $0.96$ & Y \\
                                   & Furosemide             & Aneurysm of unspecified site                & $0.92$ & Y \\
                                   & Desonide               & Chlamydial lymphogranuloma (venereum)       & $0.92$ & N \\
\midrule                                   
\multirow{5}{*}{\makecell{FR-Identity}} & Latanoprost               & Superficial injury of ankle, foot and toes & $0.98$ & N \\
                                    & Elixophyllin               & Pneumonia in diseases classified elsewhere & $0.94$ & Y \\
                                    & Furosemide                 & Aneurysm of unspecified site               & $0.92$ & Y \\
                                    & Oxistat                    & Mycosis fungoides                          & $0.90$ & Y \\
                                    & Trifluridine               & Congenital Pneumonia                       & $0.90$ & N \\
\midrule
\multirow{5}{*}{\makecell{FR-Linear}} & Kenalog                    & Unspecified contact dermatitis & $0.96$ & Y \\
                                   & Kenalog                    & Pemphigus                      & $0.96$ & Y \\
                                   & Methyprednisolone Acetate  & Nephrotic Syndrome             & $0.96$ & Y \\
                                   & Furosemide                 & Aneurysm of unspecified site   & $0.94$ & Y \\
                                   & Dexamethasone              & Pemphigus                      & $0.90$ & Y \\
\midrule
\multirow{5}{*}{\makecell{FR-Neural}} & Onglyza               & Type 2 diabetes mellitus                            & $0.98$ & Y \\
                                   & Pradaxa               & Essential (primary) hypertension                    & $0.96$ & Y \\
                                   & Oxytocin              & Pauciarticular juvenile rheumatoid arthritis        & $0.94$ & Y \\
                                   & Terbutaline sulfate   & HIV 2 as the cause of diseases classified elsewhere & $0.94$ & N \\
                                   & Lipitor               & Cerebral infarction                                 & $0.92$ & Y \\
\bottomrule
\end{tabular}
\end{adjustbox}
}
\caption{Highest confidence drug targets that were not annotated in the Roam Health Knowledge Graph.}
\label{tab:retargets_full}
\end{table*}

\begin{table}[ht]
    \centering
    \begin{tabular}{l c}
        \toprule
        \multirow{2}{2cm}{\makecell{\bf{Retrofitting}\\\bf{Model}}} & \bf{`Treats'} \\
                                                                    & (9152/2490)  \\
        \midrule
        None                                                        & $72.02\pm0.50$             \\
        FR-Identity                                                    & $72.93\pm 0.82$             \\
        FR-Linear                                                      & $\boldsymbol{84.22\pm0.82}$        \\
        FR-Neural                                                      & $\underline{73.52\pm0.89}$             
        \\
        \bottomrule
    \end{tabular}
    \caption{Drug-Disease Link Prediction Accuracies.}
    \label{tab:rhkg}
\end{table}

We also investigate the predictions induced by the retrofitted representations. An interesting use of healthcare knowledge graphs is to predict drug \textit{retargets}, that is, diseases for which there is no annotated treatment relationship with the drug but such a relationship may exist medically. As shown in Table \ref{tab:retargets_full}, the top retargets predicted by the linear retrofitting model are all medically plausible. In particular, the model confidently predicts that Kenalog would treat contact dermatitis, an effect also found in a clinical trial \cite{usatine2010diagnosis}.
The second most confident prediction of drug retargets was that Kenalog can treat pemphigus, which is indicated on Kenalog's drug label,\footnote{\url{https://www.accessdata.fda.gov/drugsatfda_docs/label/2014/014901s042lbledt.pdf}} 
but was not previously included in the knowledge graph.
The third prediction was that methyprednisolone acetate would treat nephrotic syndrome, which is reasonable as the drug is now labelled to treat idiopathic nephrotic syndrome.\footnote{\url{https://dailymed.nlm.nih.gov/dailymed/fda/fdaDrugXsl.cfm?setid=978b8416-2e88-4816-8a37-bb20b9af4b1d}}
Interestingly, several models predict that furosemide treats ``aneurysm of unspecified site", a relationship not indicated on the drug label\footnote{\url{https://dailymed.nlm.nih.gov/dailymed/drugInfo.cfm?setid=eadfe464-720b-4dcd-a0d8-45dba706bd33}}, though furosemide has been observed to reduce intracranial pressure \cite{samson1982furosemide}, a key factor in brain aneurysms.
Finally, both the distributional data and the embeddings produced by the baseline identity retrofitting model make the nonsensical prediction that Latanoprost, a medication used to treat intraocular pressure, would also treat superficial ankle and foot injuries.

The accuracy of the predictions from the more complex models underscores the utility of the new framework for retrofitting distributional embeddings to knowledge graphs with relations that do not imply similarity.

\section{Conclusions and Future Work}
We have presented \textit{Functional Retrofitting}, a new framework for \textit{post-hoc} retrofitting of entity embeddings to the structure of a knowledge graph. By explicitly modeling pairwise relations, this framework allows users to encode, learn, and extract information about relation semantics while simultaneously updating entity representations. This framework extends the popular concept of retrofitting to knowledge graphs with diverse entity and relation types. Functional Retrofitting is especially beneficial for graphs in which distinct distributional corpora are available for different entity classes, but it loses no accuracy when applied to simpler knowledge graphs. Finally, we are interested in the possibility of improvements to the optimization procedure outlined in this paper, including dynamic updates of the $\beta$ and $\alpha$ parameters to increase trust in the graph structure while the relation functions are learned.





\section*{Acknowledgements}
We would like to thank Adam Foster, Ben Peloquin, JJ Plecs, and Will Hamilton for insightful comments, and anonymous reviewers for constructive criticism.

\bibliography{retrofitting_coling}
\bibliographystyle{acl}

\appendix{}
\section{Structure of Knowledge Graphs}
\subsection{FrameNet}
\label{sec:framenet_structure}
The structure of the FrameNet \cite{baker1998berkeley,fillmore2003background} knowledge graph is shown in Table~\ref{tab:framenet_structure}.

\begin{table}[ht]
\centering
\begin{tabular}{l r r}
    \toprule
    \bf{Entity Type} & \bf{Count} & \bf{W2V Count} \\
    \midrule
    Token & 13572 & 12167 \\    
    Frame & 1221 & 464 \\
    \bottomrule
    \end{tabular}
    
    \vspace{8pt}
    
    \begin{tabular}{l r@{\ $\rightarrow$\ }l r }
        \toprule
        \bf{Edge Type} &\multicolumn{2}{c}{\bf{Connects}} & \bf{Count}\\
        \midrule
        Frame & Token & Frame & 13572 \\
        Lexical\_Unit & Frame & Token & 13572 \\
        Inheritance & Frame & Frame & 1562 \\
        Using & Frame & Frame & 1110 \\
        ReFraming\_Mapping & Frame & Frame & 428 \\
        Persepctive\_on & Frame & Frame & 242 \\
        Precedes & Frame & Frame & 178 \\
        See\_also & Frame & Frame & 172 \\
        Causative\_of & Frame & Frame & 120 \\
        Inchoative\_of & Frame & Frame & 38 \\
        Metaphor & Frame & Frame & 8 \\
        \bottomrule
    \end{tabular}
    \caption{Structure of the FrameNet knowledge graph.}
    \label{tab:framenet_structure}
\end{table}

\clearpage
\subsection{WordNet}
\label{sec:wordnet_structure}
The structure of the WordNet knowledge graph \cite{miller1995wordnet} is shown in Table~\ref{tab:wordnet_structure}.

\begin{table}[ht]
    \centering
    \begin{tabular}{l r r}
        \toprule
        \bf{Entity Type} & \bf{Count} & \bf{W2V Count} \\
        \midrule
        Lemma & 206978 & 115635 \\
        \bottomrule
    \end{tabular}
    \vspace{8pt}
    
    \begin{tabular}{l r}
        \toprule
        \bf{Edge Type} & \bf{Count}\\
        \midrule
        Hypernym & 136,235 \\
        Hyponym & 136,235 \\
        Derivationally Related Form & 60,250 \\
        Antonym & 5,922 \\
        Pertainym & 5,573 \\
        Usage Domain & 69 \\
        \bottomrule
    \end{tabular}
    \caption{Structure of the WordNet knowledge graph.}
    \label{tab:wordnet_structure}
\end{table}

\subsection{SNOMED-CT}
\label{sec:snomed_structure}
The structure of the knowledge graph extracted from the SNOMED-CT ontology is shown in Table~\ref{tab:snomed_structure}.

\begin{table}[ht]
	{\footnotesize
    \centering
    \begin{adjustbox}{center}
    \begin{tabular}{lr | lr}
        \toprule
        \bf{Edge Type} & \bf{Count} & \bf{Edge Type} & \bf{Count} \\
        \midrule
associated\_clinical\_finding&493258&child&242130\\
has\_finding\_site&205263&has\_method&200507\\
has\_associated\_morphology&169778&has\_procedure\_site&79171\\
has\_causative\_agent&69284&interprets&67900\\
has\_active\_ingredient&58976&part\_of&47776\\
has\_direct\_procedure\_site&45693&mapped\_to&37287\\
same\_as&30670&has\_pathological\_process&23641\\
has\_dose\_form&23259&has\_intent&22845\\
causative\_agent\_of&19833&finding\_site\_of&19525\\
has\_direct\_morphology&18380&has\_direct\_substance&16913\\
has\_component&15597&has\_indirect\_procedure\_site&15596\\
occurs\_in&14003&possibly\_equivalent\_to&13459\\
has\_finding\_method&12754&active\_ingredient\_of&12423\\
has\_definitional\_manifestation&11788&has\_direct\_device&11223\\
is\_interpreted\_by&10908&has\_interpretation&10077\\
procedure\_site\_of&9559&occurs\_after&7825\\
has\_temporal\_context&7786&associated\_morphology\_of&7524\\
has\_subject\_relationship\_context&7465&has\_part&6851\\
uses\_device&6407&associated\_with&6399\\
has\_measured\_component&6353&uses&6221\\
has\_associated\_finding&6205&has\_focus&6122\\
uses\_substance&5474&component\_of&5256\\
temporally\_follows&5029&due\_to&4884\\
has\_finding\_context&4883&direct\_procedure\_site\_of&4252\\
has\_specimen&3767&replaces&3726\\
has\_laterality&3641&associated\_finding\_of&3432\\
has\_associated\_procedure&3397&has\_clinical\_course&3309\\
has\_course&3241&has\_procedure\_context&2945\\
has\_approach&2808&measured\_component\_of&2741\\
has\_access&2660&has\_specimen\_source\_topography&2457\\
has\_finding\_informer&2229&has\_onset&2168\\
has\_priority&1854&mth\_xml\_form\_of&1794\\
mth\_plain\_text\_form\_of&1794&mth\_has\_xml\_form&1794\\
mth\_has\_plain\_text\_form&1794&direct\_substance\_of&1783\\
focus\_of&1680&indirect\_procedure\_site\_of&1662\\
has\_revision\_status&1599&uses\_access\_device&1587\\
has\_access\_instrument&1518&direct\_device\_of&1434\\
has\_indirect\_morphology&1426&associated\_procedure\_of&1320\\
has\_specimen\_procedure&1309&has\_communication\_with\_wound&1155\\
cause\_of&1121&has\_extent&1082\\
has\_specimen\_substance&1030&method\_of&921\\
has\_procedure\_device&770&uses\_energy&753\\
has\_procedure\_morphology&752&has\_surgical\_approach&697\\
dose\_form\_of&676&direct\_morphology\_of&673\\
referred\_to\_by&667&has\_associated\_etiologic\_finding&656\\
used\_by&608&priority\_of&586\\
specimen\_source\_topography\_of&584&occurs\_before&574\\
specimen\_procedure\_of&555&has\_severity&525\\
device\_used\_by&525&substance\_used\_by&507\\
definitional\_manifestation\_of&436&temporally\_followed\_by&406\\
has\_specimen\_source\_identity&327&has\_property&282\\
has\_instrumentation&274&has\_subject\_of\_information&272\\
has\_specimen\_source\_morphology&251&access\_instrument\_of&226\\
has\_scale\_type&206&specimen\_substance\_of&171\\
has\_episodicity&168&has\_route\_of\_administration&143\\
has\_recipient\_category&143&associated\_etiologic\_finding\_of&143\\
specimen\_of&134&approach\_of&125\\
subject\_relationship\_context\_of&115&has\_indirect\_device&114\\
interpretation\_of&109&procedure\_device\_of&107\\
course\_of&106&indirect\_morphology\_of&10\\
        \bottomrule
    \end{tabular}
    \end{adjustbox}
    \caption{Structure of the SNOMED-CT knowledge graph.}
    \label{tab:snomed_structure}
    }
\end{table}

\subsection{Roam Health Knowledge Graph}
The structure of the extracted subgraph of the RHKG is summarized in Table~\ref{tab:rhkg_structure}. A disjoint set of 11,306 drug--disease relations is reserved for evaluation.

\begin{table}[htb]
    \centering
    \begin{tabular}{l r}
        \toprule
        \bf{Entity Type}   & \bf{Count} \\
        \midrule
        Drug                      & 223,019  \\
        Disease                   & 95,559 \\
        \bottomrule
    \end{tabular}
    \vspace{8pt}
    \begin{tabular}{l r@{ \ $\rightarrow$ \ }l r}
        \toprule
        \bf{Edge Type} &\multicolumn{2}{c}{\bf{Connects}} & \bf{Count}\\
        \midrule
        Ingredient Of         & Drug    & Drug    & 49,218 \\
        Has Ingredient        & Drug    & Drug    & 49,208 \\
        Is A                  & Drug    & Drug    & 28,297 \\
        Has Descendent        & Disease & Disease & 22,344 \\
        Treats                & Drug    & Disease & 19,374 \\
        Has Active Ingredient & Drug    & Drug    & 18,422 \\
        Has Child             & Disease & Disease & 18,066 \\
        Active Ingredient Of  & Drug    & Drug    & 17,175 \\
        Has TradeName         & Drug    & Drug    & 11,783 \\
        TradeName Of          & Drug    & Drug    & 11,783 \\
        Inverse Is A          & Drug    & Drug    & 10,369 \\
        Has Symptom           & Disease & Disease & 7,892  \\
        Part Of               & Drug    & Drug    & 6,882  \\
        Has Part              & Drug    & Drug    & 6,624  \\
        Same As               & Drug    & Drug    & 5,882  \\
        Precise Ingredient Of & Drug    & Drug    & 3,562  \\
        Has Precise Ingredient& Drug    & Drug    & 3,562  \\
        Possibly Equivalent To& Drug    & Drug    & 1,233  \\
        Causative Agent of    & Drug    & Drug    & 1,070  \\
        Has Form              & Drug    & Drug    & 602    \\
        Form of               & Drug    & Drug    & 602    \\
        Component of          & Drug    & Drug    & 436    \\
        Includes              & Disease & Disease & 347    \\
        Has Dose Form         & Drug    & Drug    & 138  \\
        \bottomrule
    \end{tabular}
    \caption{Structure of the subgraph of the Roam Health Knowledge Graph.}
    \label{tab:rhkg_structure}
\end{table}

\end{document}